\documentclass{svproc}
\usepackage{makeidx}  
\makeindex
\usepackage{url}
\usepackage{amsmath}
\usepackage{amssymb}
\usepackage{graphicx}
\usepackage{array}
\usepackage{subcaption}
\usepackage{float}

\begin{document}
\title{SemGeoNav: A Safety-Guided Visual Navigation Approach with Semantic Reasoning and Geometric Planning}
\titlerunning{SemGeoNav}  
%
\author{Yu Liu\inst{1} \and  Zongyang Chen \inst{1} \and Yan Guo \inst{1} \and  Chao Liu\inst{1} \and Xianfei Pan\inst{1, 2}}
\authorrunning{Y. Liu et al.} 
%
\tocauthor{Yu Liu, Zongyang Chen, Yan Guo, Chao Liu, Xianfei Pan}
\index{Liu, Y.}
\index{Chen, Z.}
\index{Guo, Y.}
\index{Liu, C.}
\index{Pan, X.}
\institute{College of Intelligence Science and Technology, National University of Defense Technology, Changsha 410073, China.
\and
\email{afeipan@126.com}}

\maketitle              

\begin{abstract}
Learning-based visual navigation has enhanced semantic goal-reaching capabilities. However, due to their black-box nature, purely end-to-end models often lack explicit geometric constraints, leading to unpredictable and unreliable obstacle avoidance in open environments.
Conversely, traditional geometric planners ensure safety but struggle with high-dimensional visual targets.
To address these limitations, we propose SemGeoNav, a novel hierarchical visual navigation framework.
It tightly integrates the high-level semantic reasoning of end-to-end models with the reliable local planning ability of geometry-based methods, achieving robust image-based navigation while significantly improving obstacle avoidance.
Furthermore, we introduce a temporal trajectory smoothing mechanism to ensure continuous and stable robot motion.
We evaluated SemGeoNav on a Unitree Go2 quadruped robot in real-world environments. The results demonstrate that SemGeoNav outperforms existing representative methods, including ViNT and NoMaD, achieving higher success rates and shorter navigation times.

\end{abstract}
\section{Introduction}
Visual navigation has recently undergone a paradigm shift with the rapid development of learning-based methods, giving rise to a series of end-to-end general-purpose navigation models. Representative approaches such as GNM \cite{gnm}, ViNT \cite{vint}, and NoMaD \cite{nomad} have demonstrated impressive zero-shot generalization capabilities. By directly mapping current RGB observation sequences and goal images to motion commands, these methods enable robots to navigate in mapless environments. However, purely end-to-end models lack explicit geometric reasoning and typically rely on implicitly predicted waypoints or latent policies for obstacle avoidance. As a result, their safety behavior is largely learned from large-scale training data rather than enforced by physical or spatial constraints. When deployed on unseen robotic platforms or in the presence of unfamiliar obstacles, such methods may exhibit unreliable avoidance behavior, leading to an increased risk of collision in real-world settings.

Conversely, traditional geometry-based navigation methods, such as the A* algorithm \cite{astar} and the Dynamic Window Approach (DWA) \cite{dwa}, rely on explicit geometric representations and often require pre-built maps for path planning and obstacle avoidance. More recent local geometric planners, such as Ego-Planner \cite{egoplanner}, relax this requirement by constructing local environment representations online and optimizing collision-free trajectories in real time without a global map. Although these methods provide more reliable and intuitive obstacle avoidance, they generally suffer from limited generalization and lack the ability to understand or reason about semantic goals in the environment. 

To integrate semantic reasoning and geometric obstacle avoidance, recent research has increasingly explored hybrid frameworks that combine learning-based navigation with geometric planning. One representative direction uses classical geometric planners to provide geometric supervision for training visual foundation models (VFMs), thereby improving the robustness of the learned navigation policy \cite{lessismore}. However, in this representative line of work, geometry mainly influences the policy before inference, while deployment still depends on the implicit obstacle-avoidance behavior learned by the VFM. In contrast, SemGeoNav adds an explicit geometric planning module after the implicit VFM-based inference stage. The VFM first generates semantically guided motion candidates, and the geometric module further optimizes these trajectories based on depth observations to improve obstacle avoidance. This design preserves the semantic goal-reaching ability of the VFM while explicitly improving local geometric safety during inference.

To address these challenges, we propose SemGeoNav, a hierarchical navigation framework that preserves high-level semantic reasoning while enhancing explicit geometry-aware obstacle avoidance. 
Specifically, our framework leverages an end-to-end network to extract global contextual features, which are passed to a diffusion model for generating diverse candidate trajectories and preserving high-level semantic reasoning. To improve safety, we introduce a TTC-based risk estimation module to proactively identify and suppress high-risk commands at an early stage. These candidate trajectories are further refined by a gradient-based optimizer to obtain safer trajectories. Finally, a temporal smoothing mechanism is applied to improve motion continuity and reduce execution jitter during real-world robot deployment.
Extensive real-world experiments demonstrate that our method achieves safer and more stable navigation performance in cluttered environments.

Our contributions are summarized as follows:
\begin{itemize}
\item We propose SemGeoNav, a hierarchical navigation framework that seamlessly integrates semantic and geometric information. Our method explicitly accounts for geometric and physical constraints while preserving robust global semantic reasoning. 
\item We develop a real-time geometric planner that leverages depth images to refine the navigation commands generated by an end-to-end model, significantly improving obstacle avoidance and success rates in unstructured environments.
\item We collect a dedicated dataset on the Unitree Go2 quadruped platform and conduct targeted training for the proposed method, thereby improving its navigation performance and robustness in real-world deployment.
\end{itemize}

\begin{figure}[H]
    \centering
    \includegraphics[trim=2.0cm 0.5cm 2.0cm 0.5cm, clip, width=1.0\textwidth]{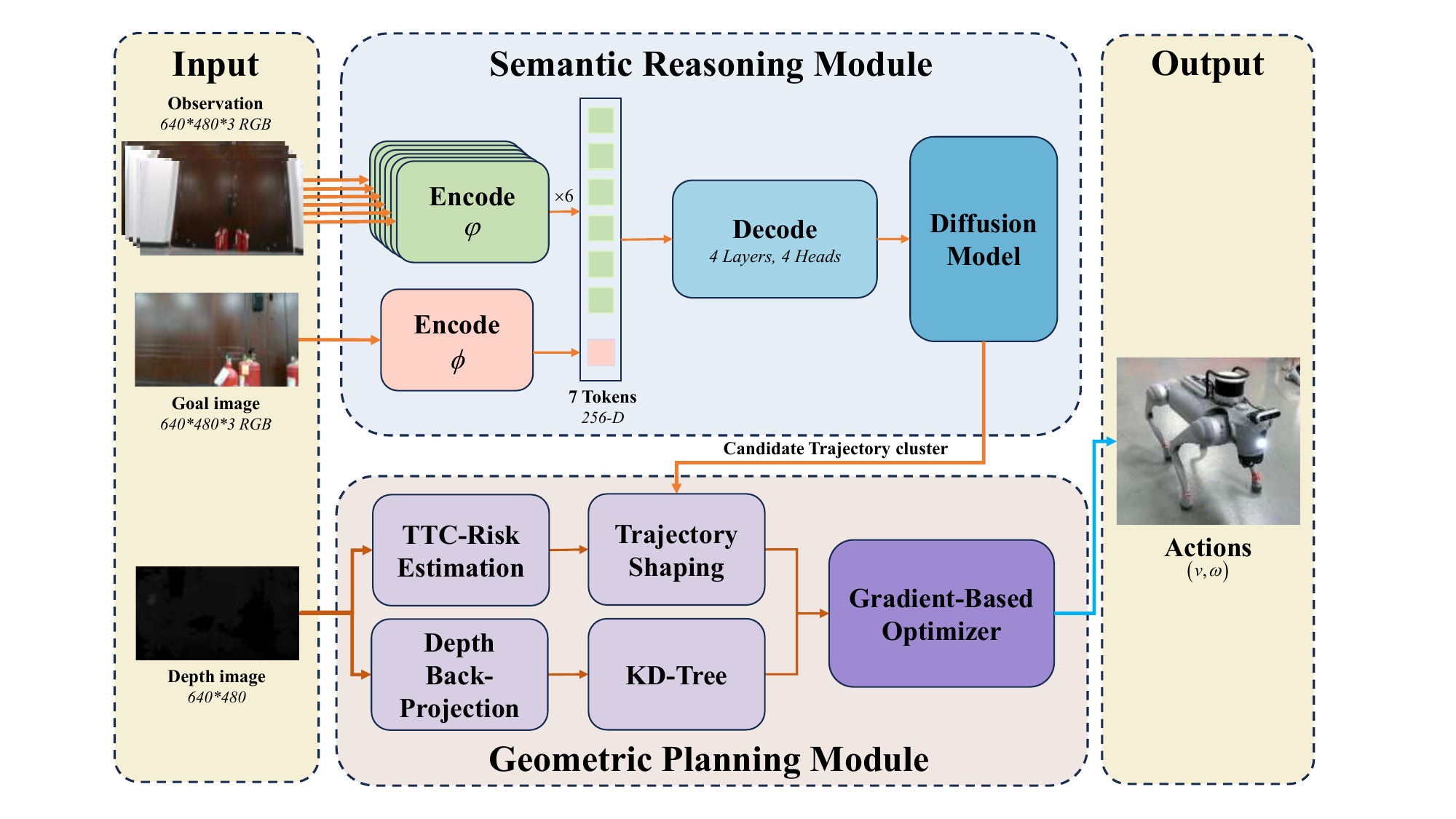}
    \caption{Overview of our proposed SemGeoNav framework.}
    \label{fig:icgnc}
\end{figure}

\section{Related Work}

\subsection{End-to-End Visual Navigation}

Learning-based visual navigation is undergoing a paradigm shift from task-specific policies to more generalizable foundation models. Early systems such as BADGR \cite{badgr} demonstrated the potential of end-to-end navigation in unstructured environments. Building on this foundation, large-scale visual foundation models for embodied agents \cite{embodiedintelligence} were introduced to improve generalization across tasks and platforms.
Meanwhile, another line of research has developed general topological navigation architectures. GNM \cite{gnm} pioneered cross-embodiment training. ViNT \cite{vint} advanced this paradigm with efficient Transformer-based architectures. NoMaD \cite{nomad} further incorporated diffusion models to capture multimodal navigation behaviors and improve exploration in unseen environments.
Nevertheless, despite their strong goal-conditioned semantic reasoning capabilities, these methods still lack explicit geometric grounding and strict safety guarantees during real-world deployment.

\subsection{Traditional Geometry-Based Navigation}

Traditional geometry-based navigation methods prioritize geometric safety and physical feasibility. These include global search algorithms like A* \cite{astar}, as well as reactive local planners like DWA \cite{dwa} and TEB \cite{teb}. All of these methods can achieve reliable obstacle avoidance when an accurate map is available.
To reduce the high computational overhead of building dense maps, gradient-based trajectory optimization methods such as Fast-Planner \cite{fastplanner} and Ego-Planner \cite{egoplanner} have been proposed. These methods do not require a pre-built global map and can generate smooth and safe trajectories by constructing local maps in real time.
However, these methods are inherently semantically blind. They treat all point clouds as rigid obstacles and cannot reason about traversable terrain or high-level goals.

\subsection{Hybrid Visual-Geometric Navigation}

Recent navigation methods have started to integrate semantic understanding with geometric safety. Representative approaches include Less Is More \cite{lessismore}, LM-Nav \cite{lmnav}, and Embodied-R1 \cite{embodiedr1}. These methods use visual models to generate spatial sub-goals that are not constrained by a dedicated geometric module, and then pass them to geometric controllers for execution. 
However, most of these methods use geometry mainly for sub-goal execution or expect obstacle avoidance to be implicitly learned from data, rather than tightly coupling semantic intent and geometric safety during online command generation.

\section{Methodology}

As depicted in Fig.~\ref{fig:icgnc}, SemGeoNav tightly integrates high-level semantic reasoning with low-level geometric planning. First, the current observations and goal image are encoded into a contextual representation, which is used to condition a diffusion model for candidate trajectory generation. Next, the geometric planning module performs depth back-projection to construct a local obstacle representation and applies TTC-based risk estimation for early safety filtering. The filtered trajectories are then refined by a gradient-based optimizer under geometric safety constraints. Finally, temporal smoothing is applied to produce a safe and stable control command.

\subsection{Semantic Reasoning Module}

Following ViNT \cite{vint} and NoMaD \cite{nomad}, we employ a hybrid architecture that combines EfficientNet \cite{efficientnet} and a Transformer \cite{transformer}. EfficientNet enhances the local feature extraction capabilities, improving robustness across diverse scenes. The Vision Transformer further captures global feature associations through its self-attention mechanism. 
When the current observation sequence and the target image are input into our Semantic Reasoning Module, both the observation images and the goal image are encoded using EfficientNet-B0.
Define $o_{t-P+1:t}=\{o_{t-P+1},\ldots,o_t\}$ as the current RGB observation sequence and $o_g$ as the target image. Let $\psi(\cdot)$ denote the visual encoder and $\phi(\cdot)$ denote the target encoder. The visual encoding $z_i$ obtained through EfficientNet and the target encoding $z_g$ are defined as:
\begin{equation}
z_i = \psi(o_i), \quad i \in \{t-P+1,\ldots,t\}; \qquad z_g = \phi(o_g)
\end{equation}

The resulting encodings are then treated as input tokens and aggregated by a Transformer backbone $F_\theta$ to produce a semantic context vector:
\begin{equation}
c_t = F_\theta\big([z_{t-P+1},\ldots,z_t,z_g]\big)
\end{equation}

The semantic context $c_t$ summarizes the current scene, the recent motion history, and the relation between the current view and the goal. This enables the generated navigation commands to reason about future motion intentions rather than instantaneous control commands. 
In contrast to NoMaD, which introduces additional exploration strategies for non-goal-oriented navigation, our method considers only goal-oriented navigation within the visible range. As a result, the goal mask mechanism is omitted.

On top of the shared context $c_t$, we model the navigation policy as a diffusion model \cite{ddpm} over a short-horizon waypoint sequence $\mathbf{w}_t$. Rather than predicting a single deterministic command, the diffusion head learns a conditional distribution over future waypoints:
\begin{equation}
p(\mathbf{w}_t \mid c_t)
\end{equation}

At inference time, we initialize the waypoint sequence with Gaussian noise, $\mathbf{w}_t^{K}\sim\mathcal{N}(0,I)$, and iteratively denoise it for $K$ steps:
\begin{equation}
\mathbf{w}_t^{k-1}=\alpha_k\!\left(\mathbf{w}_t^{k}-\gamma_k\,\epsilon_\theta(c_t,\mathbf{w}_t^{k},k)\right)+\sigma_k \eta,
\end{equation}

where $\epsilon_\theta(c_t,\mathbf{w}_t^{k},k)$ is the noise predicted by the diffusion network at denoising step $k$, $\alpha_k$ is a step-dependent scaling coefficient, $\gamma_k$ controls the update magnitude during denoising, $\sigma_k$ determines the noise level injected at step $k$, and $\eta \sim \mathcal{N}(0,I)$ is standard Gaussian noise. The final output $\mathbf{w}_t^{0}$ is the predicted short-horizon waypoint sequence conditioned on the semantic context $c_t$.

This semantic strategy is trained through an integrated objective function, which includes the diffusion denoising loss and the auxiliary distance prediction loss:
\begin{equation}
\mathcal{L}_{\text{sem}}=\mathbb{E}\Big[\|\eta-\epsilon_\theta(c_t,\mathbf{w}_t^{k},k)\|_2^2+\lambda_d\,\ell_d(\hat d_t,d_t)\Big]
\end{equation}

Here, $\eta$ is the Gaussian noise added to the clean waypoint sequence during the forward diffusion process, $\hat d_t$ is the predicted temporal distance to the goal, $d_t$ is the corresponding supervision signal, $\ell_d(\cdot,\cdot)$ is the distance prediction loss function, and $\lambda_d$ is a weighting coefficient that balances the two objectives. The denoising term enables the model to learn a multimodal distribution over feasible future waypoints, while the auxiliary distance prediction term encourages the semantic module to capture goal progress information.

\subsection{Geometric Planning Module}

Given the semantic waypoint candidates produced by the high-level policy, we employ a geometry-aware local planner to convert them into safe and executable control commands. The planner consists of four components: local geometric representation from depth, TTC-based risk shaping, collision-aware trajectory refinement, and final trajectory scoring with temporal smoothing. In this way, the semantic policy provides the motion intent, while the geometric module enforces collision safety and execution stability.

\paragraph{Depth Back-Projection: }
We first convert the depth image into an explicit local obstacle representation in the robot frame. Given a depth map $D_t \in \mathbb{R}^{H \times W}$, we back-project valid pixels into a set of 2D obstacle points using the camera intrinsics $(f_x,f_y,c_x,c_y)$. Since collision checking is performed on the local ground plane, we retain only the forward and lateral coordinates in the robot frame. For a pixel coordinate $(u,v)$ with depth value $z_{(u,v)}$, the corresponding 2D point is approximated by
\begin{equation}
x_{(u,v)} = z_{(u,v)}, \qquad
y_{(u,v)} = -\frac{(u-c_x)\,z_{(u,v)}}{f_x}.
\end{equation}

Here, $x_{(u,v)}$ and $y_{(u,v)}$ denote the forward and lateral coordinates, respectively. After subsampling and filtering invalid measurements, we obtain a compact local obstacle set
\begin{equation}
\mathcal{P}_t = \{p_j\}_{j=1}^{N}, \qquad p_j = [x_j, y_j]^\top,
\end{equation}

where $N$ is the number of retained obstacle points, and $p_j$ denotes the $j$-th obstacle point in the local obstacle set. To accelerate collision queries, we build a KD-tree over $\mathcal{P}_t$.

\paragraph{TTC-Based Risk Estimation: }
Before the subsequent gradient-based optimization and trajectory selection, we first estimate the short-term collision risk from the depth image. In our implementation, we divide the image into a central region and two side regions, and compute a robust depth statistic for each region. Let $z_t^{\mathrm{c}}$ denote the first decile of depth in the center region at time $t$, and let $z_{t-1}^{\mathrm{c}}$ denote the first decile of depth in the center region at time $t-1$. If the center depth decreases over time, the approach speed of the robot toward the obstacle ahead can be approximated as
\begin{equation}
v_t=\frac{z_{t-1}^{\mathrm{c}}-z_t^{\mathrm{c}}}{\Delta t},
\end{equation}

where $\Delta t$ denotes the time interval between two consecutive frames. The corresponding time-to-contact can then be approximated as
\begin{equation}
\mathrm{TTC}_t=\frac{z_t^{\mathrm{c}}}{v_t}, \qquad v_t>0.
\end{equation}

When $\mathrm{TTC}_t$ falls below a predefined threshold $\tau_{\mathrm{ttc}}$, or when the center depth $z_t^{\mathrm{c}}$ becomes smaller than the safety distance $d_{\mathrm{safe}}$, the planner enters an emergency mode. In this mode, the high-risk components of the candidate motion are strongly suppressed, and the control command is temporally smoothed over a short horizon to avoid abrupt oscillatory behavior. Furthermore, when the traversable space in a certain direction becomes smaller, the motion component along that direction is also suppressed. In this way, the TTC-based risk estimation module acts as a fast safety filter that reshapes high-risk candidates before the subsequent fine-grained trajectory optimization.

The empirical thresholds are selected according to the robot footprint, maximum velocity, control frequency, and depth noise. This module is used as a lightweight front-end safety filter, rather than an independent optimal planner. The TTC threshold $\tau_{\mathrm{ttc}}$ and the safety distance $d_{\mathrm{safe}}$ mainly determine when the emergency mode is activated. Smaller values delay the response to frontal obstacles, while larger values trigger earlier intervention and make the robot more conservative. The side-distance threshold $d_{\mathrm{s}}$ controls the suppression of lateral motion when the traversable space on one side becomes limited. The smoothing factor $\beta$ adjusts the temporal smoothness of the output command and helps reduce abrupt oscillations. In validation, moderate perturbations of these parameters did not change the overall navigation trend. They mainly affected the trade-off between safety margin, motion smoothness, and conservativeness in narrow passages.

\paragraph{Gradient-Based Optimizer: }
Let $\mathbf{q}=\{q_1,\ldots,q_M\}$ denote a candidate short-horizon trajectory predicted by the semantic policy, where $q_i \in \mathbb{R}^2$ denotes the $i$-th waypoint on the predicted trajectory, and $M$ is the total number of waypoints. 
We refine each candidate trajectory and score it with a unified objective:
\begin{equation}
C(\mathbf{q}) = \alpha_c J_{\mathrm{c}}(\mathbf{q}) + \alpha_s J_{\mathrm{s}}(\mathbf{q}) + \alpha_d J_{\mathrm{d}}(\mathbf{q})
- \alpha_p R_{\mathrm{p}}(\mathbf{q}) - \alpha_r R_{\mathrm{c}}(\mathbf{q}),
\end{equation}

where $\alpha_c$, $\alpha_s$, $\alpha_d$, $\alpha_p$, and $\alpha_r$ are weighting coefficients. Here, $J_{\mathrm{d}}(\mathbf{q})$ measures the deviation between the refined trajectory and the original trajectory, $R_{\mathrm{p}}(\mathbf{q})$ rewards forward progress toward the goal, and $R_{\mathrm{c}}(\mathbf{q})$ encourages the trajectory to stay farther away from obstacles. $J_{\mathrm{s}}(\mathbf{q})$ denotes the smoothness cost, and $J_{\mathrm{c}}(\mathbf{q})$ denotes the collision cost. They are defined as follows:
\begin{equation}
J_{\mathrm{s}}(\mathbf{q})= \sum_{i=2}^{M-1} \left\|q_{i+1} - 2q_i + q_{i-1}\right\|_2^2
\end{equation}
\begin{equation}
J_{\mathrm{c}}(\mathbf{q})= \sum_{i=1}^{M} \max(0, d_{\mathrm{s}} - \delta_i)^3
\end{equation}

Here, $d_{\mathrm{s}}$ is determined by the robot footprint and an additional safety margin, and is set to 0.25\,m in our implementation. The distance term $\delta_i$ is defined as
\begin{equation}
\delta_i = \min_{b \in \mathcal{B}(q_i)} \min_{p \in \mathcal{P}_t} \|b - p\|_2,
\end{equation}
where $\mathcal{B}(q_i)$ consists of sampled body points at waypoint $q_i$. In addition, $J_{\mathrm{d}}(\mathbf{q})$ measures the deviation from the original semantic trajectory, $R_{\mathrm{p}}(\mathbf{q})$ rewards forward progress toward the goal, and $R_{\mathrm{c}}(\mathbf{q})$ encourages larger clearance from nearby obstacles. The final trajectory is selected as
\begin{equation}
\mathbf{q}_t^\star = \arg\min_{\mathbf{q}} C(\mathbf{q}).
\end{equation}

Let $w_t$ denote the candidate waypoint selected from the best trajectory $\mathbf{q}_t^\star$. To improve execution stability, we apply a temporal smoothing filter:
\begin{equation}
\hat w_t = \beta \hat w_{t-1} + (1-\beta) w_t,
\end{equation}
where $\hat w_t$ is the smoothed waypoint command sent to the controller, and $\beta \in [0,1)$ is the smoothing coefficient. In addition, we constrain command increments between consecutive time steps to suppress jitter, especially in narrow passages.

\section{Experiments}

\subsection{Experimental Setup}

We evaluate our method on a real-world Unitree Go2 quadruped robot in indoor visual navigation tasks. 
Following the real-robot evaluation protocol adopted in NoMaD, we deploy the system in a closed-loop setting, where perception, policy inference, and motion control are executed on a physical robot platform. 
Specifically, the sensor streams provided by the Go2, including RGB observations and depth measurements, are transmitted to an external computer through a ROS master--slave communication framework. Both the RGB images and the depth images are captured at a resolution of $640 \times 480$. 
The computer is equipped with an NVIDIA GeForce RTX 4060 GPU, on which all policy inference and local planning computations are performed in real time. The predicted linear and angular velocity commands, denoted by $(v,\omega)$, are then sent back to the robot to control its motion.

To systematically evaluate the proposed method, we construct three representative navigation scenarios: 
(1) obstacle-free navigation; 
(2) navigation with a single static obstacle; 
(3) navigation with multiple static obstacles. 
\begin{figure}[H]
    \centering
    \begin{subfigure}[b]{0.32\textwidth}
        \centering
        \includegraphics[width=\textwidth,height=2.6cm,keepaspectratio]{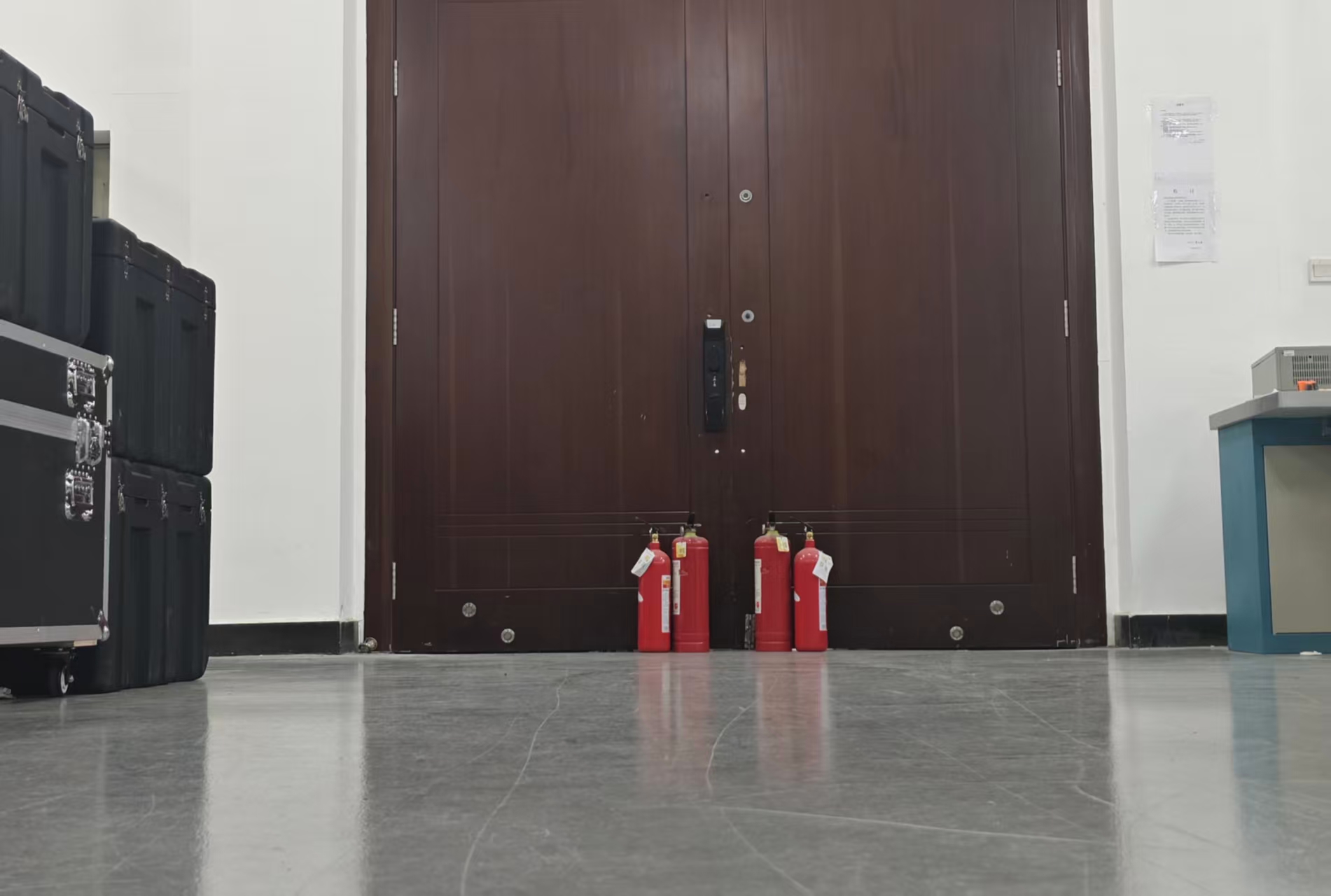}
        \caption{Obstacle-free.}
        \label{fig:free_obstacle}
    \end{subfigure}
    \hfill
    \begin{subfigure}[b]{0.32\textwidth}
        \centering
        \includegraphics[width=\textwidth,height=2.6cm,keepaspectratio]{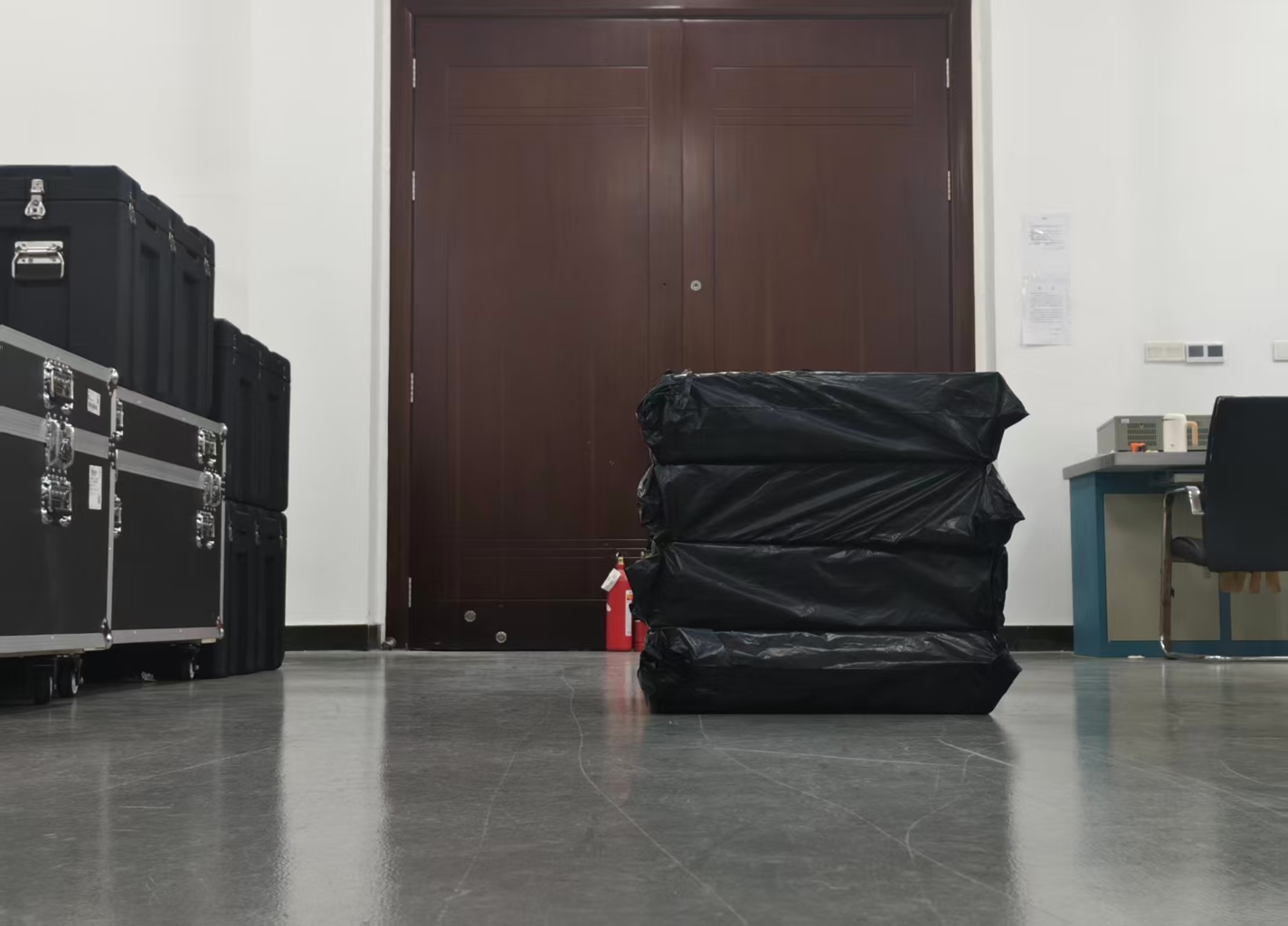}
        \caption{Single-obstacle.}
        \label{fig:single_obstacle}
    \end{subfigure}
    \hfill
    \begin{subfigure}[b]{0.32\textwidth}
        \centering
        \includegraphics[width=\textwidth,height=2.6cm,keepaspectratio]{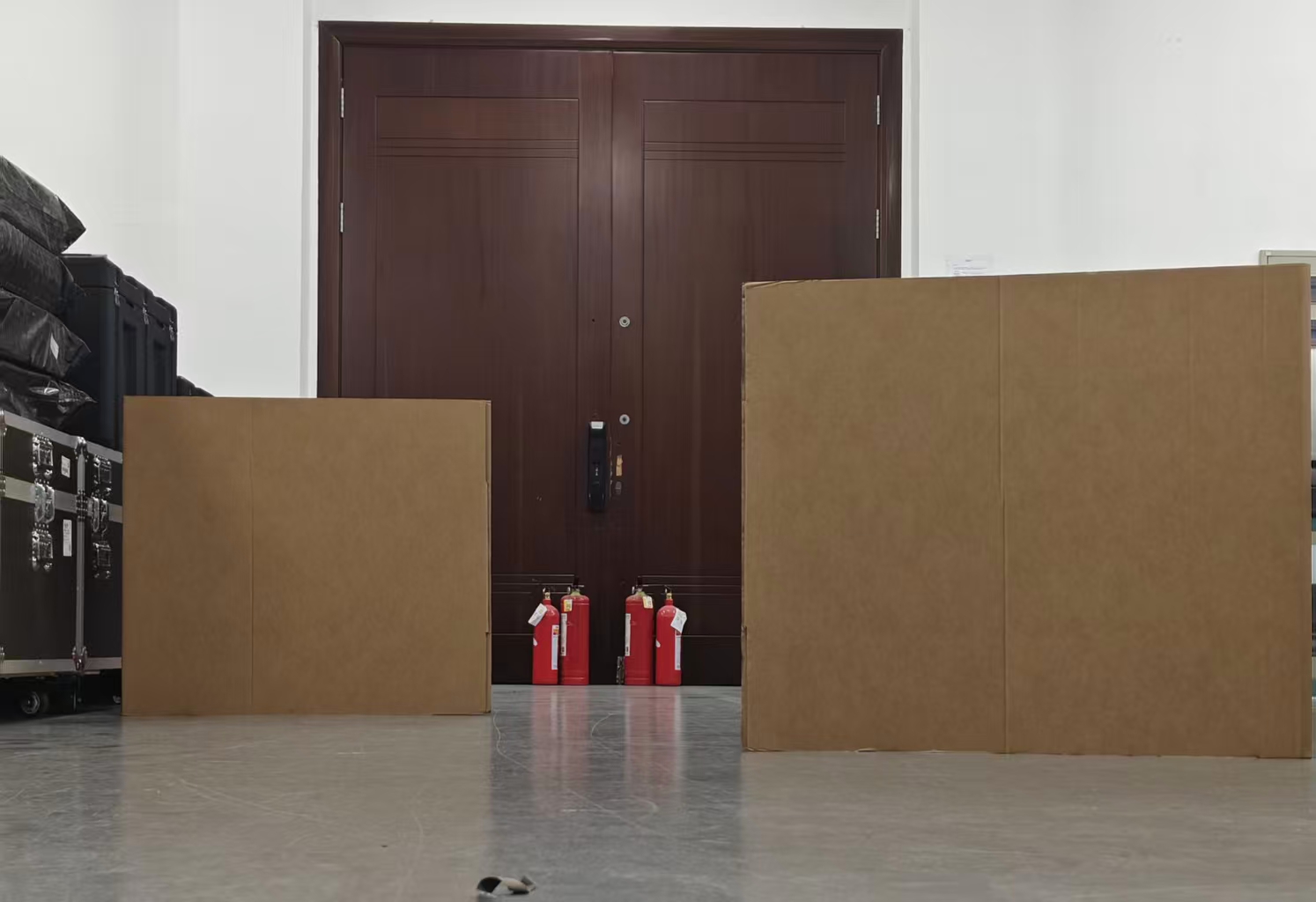}
        \caption{Multiple-obstacle.}
        \label{fig:multi_obstacle}
    \end{subfigure}
    \caption{Experimental scenes in the real-world navigation evaluation.}
    \label{fig:obstacle_scenes}
\end{figure}

For comparison, we evaluate our method against two representative vision-based navigation baselines, ViNT and NoMaD, both using the official pre-trained weights provided by the authors. 
In addition, to assess the effectiveness of the collected dataset for platform-specific adaptation, we also report results for ViNT fine-tuned on our dataset.
For fairness, all compared methods are deployed on the same Go2 platform, use the same sensor inputs and goal specifications, and share the same ROS-based control interface. All methods are evaluated using the same goal images and initial positions, with a fixed start-to-goal distance of 11 meters. 
For all methods, the output control commands are bounded by a maximum forward linear velocity of 0.4 m/s and a maximum angular velocity of 0.8 rad/s.

\subsection{Experimental Results}

The evaluation results under the obstacle-free setting are summarized in Table 1. Here, ViNT and NoMaD denote the results obtained using the official pre-trained weights provided by the authors, while ViNT-Go2 denotes the ViNT model evaluated with our fine-tuned weights. Ours denotes the proposed method. SI denotes the success indicator for each trial, where 1 means success and 0 means failure. Time denotes the travel duration from the start point to the goal.

\vspace{-2.0em}
\begin{table}[H]
\centering
\caption{Experimental results of different navigation methods in the obstacle-free scenario.}
\scriptsize
\renewcommand{\arraystretch}{1.0}
\setlength{\tabcolsep}{2pt}

\begin{tabular}{
>{\centering\arraybackslash}p{0.18\textwidth}|
>{\centering\arraybackslash}p{0.07\textwidth}
>{\centering\arraybackslash}p{0.07\textwidth}|
>{\centering\arraybackslash}p{0.07\textwidth}
>{\centering\arraybackslash}p{0.07\textwidth}|
>{\centering\arraybackslash}p{0.07\textwidth}
>{\centering\arraybackslash}p{0.07\textwidth}|
>{\centering\arraybackslash}p{0.09\textwidth}
>{\centering\arraybackslash}p{0.09\textwidth}
}
\hline
- & \multicolumn{2}{c|}{ViNT} & \multicolumn{2}{c|}{NoMaD} & \multicolumn{2}{c|}{ViNT-Go2} & \multicolumn{2}{c}{SemGeoNav (Ours)} \\
\hline
Sequence & SI & Time & SI & Time & SI & Time & SI & Time \\
\hline
1  & 1 & 34.06 & 1 & 30.54 & 1 & 32.07 & 1 & 29.06 \\
2  & 0 & -     & 1 & 30.00 & 1 & 31.84 & 1 & 29.31 \\
3  & 1 & 32.32 & 1 & 31.80 & 1 & 31.57 & 1 & 30.06 \\
4  & 1 & 32.32 & 1 & 31.56 & 1 & 31.07 & 1 & 26.81 \\
5  & 1 & 32.82 & 1 & 31.10 & 1 & 31.32 & 1 & 29.81 \\
6  & 1 & 33.83 & 0 & -     & 1 & 32.32 & 1 & 29.30 \\
7  & 1 & 32.58 & 1 & 31.31 & 1 & 32.57 & 1 & 31.08 \\
8  & 1 & 33.08 & 1 & 31.57 & 1 & 31.07 & 1 & 30.60 \\
9  & 1 & 33.82 & 1 & 31.56 & 1 & 30.81 & 1 & 29.83 \\
10 & 0 & -     & 1 & 30.57 & 1 & 30.06 & 1 & 30.84 \\
11 & 1 & 31.83 & 1 & 30.58 & 1 & 32.58 & 1 & 29.84 \\
12 & 1 & 31.82 & 0 & -     & 1 & 32.06 & 1 & 29.56 \\
13 & 1 & 31.82 & 0 & -     & 1 & 31.07 & 1 & 28.83 \\
14 & 1 & 31.08 & 1 & 32.56 & 1 & 29.58 & 1 & 26.35 \\
15 & 1 & 30.59 & 1 & 30.30 & 1 & 31.56 & 1 & 29.08 \\
16 & 1 & 33.07 & 1 & 31.80 & 1 & 29.82 & 1 & 28.83 \\
17 & 1 & 31.83 & 1 & 30.80 & 1 & 31.56 & 1 & 30.58 \\
18 & 1 & 30.09 & 0 & -     & 1 & 30.57 & 1 & 28.58 \\
19 & 0 & -     & 1 & 34.62 & 1 & 31.84 & 1 & 29.82 \\
20 & 1 & 32.08 & 0 & -     & 1 & 30.82 & 1 & 27.58 \\
\hline
Average time & \multicolumn{2}{c|}{32.30} & \multicolumn{2}{c|}{31.38} & \multicolumn{2}{c|}{31.31} & \multicolumn{2}{c}{\textbf{29.29}} \\
\hline
SR (success rate) & \multicolumn{2}{c|}{85\%} & \multicolumn{2}{c|}{75\%} & \multicolumn{2}{c|}{\textbf{100\%}} & \multicolumn{2}{c}{\textbf{100\%}} \\
\hline
\end{tabular}
\vspace{-0.8em}
\end{table}
\vspace{-0.8em}

Table 1 presents the experimental results in the obstacle-free scenario. Although this scenario does not require explicit obstacle avoidance, SemGeoNav still achieves the best overall performance, attaining a 100\% success rate and the shortest average navigation time of 29.29\,s. This suggests that the introduction of the geometric planning module does not compromise the goal-oriented capability of the visual navigation model, but instead improves execution stability and navigation efficiency. Moreover, after being fine-tuned on our collected dataset, ViNT improves its success rate from 85\% to 100\%, further demonstrating the effectiveness of the collected dataset for platform-specific adaptation. Overall, these results validate that both the proposed framework and the collected training data contribute to improved real-world navigation performance.

\vspace{-2.0em}
\begin{table}[H]
\centering
\caption{Experimental results of different navigation methods in the single-obstacle scenario.}
\scriptsize
\renewcommand{\arraystretch}{1.0}
\setlength{\tabcolsep}{2pt}
\begin{tabular}{
>{\centering\arraybackslash}p{0.18\textwidth}|
>{\centering\arraybackslash}p{0.07\textwidth}
>{\centering\arraybackslash}p{0.07\textwidth}|
>{\centering\arraybackslash}p{0.07\textwidth}
>{\centering\arraybackslash}p{0.07\textwidth}|
>{\centering\arraybackslash}p{0.07\textwidth}
>{\centering\arraybackslash}p{0.07\textwidth}|
>{\centering\arraybackslash}p{0.09\textwidth}
>{\centering\arraybackslash}p{0.09\textwidth}
}
\hline
- & \multicolumn{2}{c|}{ViNT} & \multicolumn{2}{c|}{NoMaD} & \multicolumn{2}{c|}{ViNT-Go2} & \multicolumn{2}{c}{SemGeoNav (Ours)} \\
\hline
Sequence & SI & Time & SI & Time & SI & Time & SI & Time \\
\hline
1  & 1 & 32.07 & 1 & 29.09 & 1 & 29.08 & 1 & 29.83 \\
2  & 1 & 31.56 & 0 & -     & 1 & 30.57 & 1 & 29.07 \\
3  & 1 & 31.32 & 1 & 30.58 & 1 & 29.36 & 1 & 28.32 \\
4  & 1 & 31.07 & 0 & -     & 1 & 31.57 & 1 & 30.09 \\
5  & 1 & 31.83 & 1 & 29.58 & 0 & -     & 1 & 30.33 \\
6  & 0 & -     & 0 & -     & 1 & 27.35 & 1 & 29.57 \\
7  & 1 & 31.32 & 0 & -     & 1 & 31.07 & 1 & 27.33 \\
8  & 1 & 31.82 & 1 & 29.08 & 1 & 27.82 & 1 & 28.34 \\
9  & 1 & 32.32 & 1 & 29.31 & 1 & 30.32 & 1 & 27.36 \\
10 & 1 & 31.57 & 0 & -     & 1 & 31.57 & 1 & 29.33 \\
\hline
Average time & \multicolumn{2}{c|}{31.65} & \multicolumn{2}{c|}{29.53} & \multicolumn{2}{c|}{29.86} & \multicolumn{2}{c}{\textbf{28.96}} \\
\hline
SR (success rate) & \multicolumn{2}{c|}{90\%} & \multicolumn{2}{c|}{50\%} & \multicolumn{2}{c|}{90\%} & \multicolumn{2}{c}{\textbf{100\%}} \\
\hline
\end{tabular}
\label{tab:single_obstacle}
\vspace{-1.5em}
\end{table}
\vspace{-1.5em}

\vspace{-2.0em}
\begin{table}[H]
\centering
\caption{Experimental results of different navigation methods in the multiple-obstacle scenario.}
\scriptsize
\renewcommand{\arraystretch}{1.0}
\setlength{\tabcolsep}{2pt}
\begin{tabular}{
>{\centering\arraybackslash}p{0.18\textwidth}|
>{\centering\arraybackslash}p{0.07\textwidth}
>{\centering\arraybackslash}p{0.07\textwidth}|
>{\centering\arraybackslash}p{0.07\textwidth}
>{\centering\arraybackslash}p{0.07\textwidth}|
>{\centering\arraybackslash}p{0.07\textwidth}
>{\centering\arraybackslash}p{0.07\textwidth}|
>{\centering\arraybackslash}p{0.09\textwidth}
>{\centering\arraybackslash}p{0.09\textwidth}
}
\hline
- & \multicolumn{2}{c|}{ViNT} & \multicolumn{2}{c|}{NoMaD} & \multicolumn{2}{c|}{ViNT-Go2} & \multicolumn{2}{c}{SemGeoNav (Ours)} \\
\hline
Sequence & SI & Time & SI & Time & SI & Time & SI & Time \\
\hline
1  & 1 & 29.82 & 0 & - & 0 & - & 1 & 26.59 \\
2  & 1 & 31.58 & 0 & - & 0 & - & 1 & 31.33 \\
3  & 0 & - & 0 & - & 1 & 30.82 & 1 & 30.94 \\
4  & 0 & - & 0 & - & 1 & 29.81 & 1 & 32.33 \\
5  & 0 & - & 0 & - & 1 & 28.82 & 1 & 27.57 \\
6  & 1 & 30.82 & 0 & - & 0 & - & 1 & 26.59 \\
7  & 0 & - & 0 & - & 1 & 30.08 & 0 & - \\
8  & 1 & 32.32 & 0 & - & 1 & 28.84 & 1 & 31.34 \\
9  & 0 & - & 0 & - & 0 & - & 0 & - \\
10 & 0 & - & 0 & - & 0 & - & 1 & 28.10 \\
\hline
Average time & \multicolumn{2}{c|}{31.14} & \multicolumn{2}{c|}{--} & \multicolumn{2}{c|}{29.67} & \multicolumn{2}{c}{\textbf{29.35}} \\
\hline
SR (success rate) & \multicolumn{2}{c|}{40\%} & \multicolumn{2}{c|}{0\%} & \multicolumn{2}{c|}{50\%} & \multicolumn{2}{c}{\textbf{80\%}} \\
\hline
\end{tabular}
\label{tab:multiple_obstacle}
\vspace{-1.0em}
\end{table}
\vspace{-1.0em}

Table 2 and Table 3 present the experimental results in environments with obstacles. When obstacles are present, the advantage of SemGeoNav becomes apparent.
This advantage becomes more pronounced in the multiple-obstacle scenario. Compared with ViNT-Go2, our method improves the success rate by 30\%. Compared with ViNT, the improvement reaches 40\%.
SemGeoNav also achieves highly competitive navigation efficiency and obtains the best average navigation time.
In contrast, the performance of NoMaD decreases significantly when obstacles are present. A possible reason is that NoMaD directly executes the first trajectory generated by the diffusion model, without explicit geometric refinement during inference.
Overall, these results show that our method is more robust and effective in obstacle-rich environments.

\subsection{Ablation Study}

We conduct an ablation study to evaluate the contribution of each component in the geometric planning module of SemGeoNav. All experiments are performed in the multiple-obstacle setting. All variants are implemented within the SemGeoNav framework and are obtained by removing or replacing specific components of its geometric planning module. 

Gradient-Only removes the TTC-based risk filtering and temporal smoothing from the geometric planning module, leaving only the gradient-based optimizer. TTC-Only removes the gradient-based refinement and uses only TTC-based risk estimation to reshape candidate motions. APF-Only replaces the geometric planner in SemGeoNav with an artificial potential field planner. SemGeoNav (Ours) is the complete framework with all proposed geometric planning components. 
The experimental results are shown in Table 4. 

\vspace{-2.0em}
\begin{table}[H]
\centering
\caption{Ablation study results of different geometric planning methods in the multiple-obstacle scenario.}
\scriptsize
\renewcommand{\arraystretch}{1.0}
\setlength{\tabcolsep}{2pt}
\begin{tabular}{
>{\centering\arraybackslash}p{0.18\textwidth}|
>{\centering\arraybackslash}p{0.07\textwidth}
>{\centering\arraybackslash}p{0.07\textwidth}|
>{\centering\arraybackslash}p{0.07\textwidth}
>{\centering\arraybackslash}p{0.07\textwidth}|
>{\centering\arraybackslash}p{0.07\textwidth}
>{\centering\arraybackslash}p{0.07\textwidth}|
>{\centering\arraybackslash}p{0.09\textwidth}
>{\centering\arraybackslash}p{0.09\textwidth}
}
\hline
- & \multicolumn{2}{c|}{Gradient-Only} & \multicolumn{2}{c|}{TTC-Only} & \multicolumn{2}{c|}{APF-Only} & \multicolumn{2}{c}{SemGeoNav (Ours)} \\
\hline
Sequence & SI & Time & SI & Time & SI & Time & SI & Time \\
\hline
1  & 0 & -     & 0 & -     & 0 & -     & 1 & 26.59 \\
2  & 1 & 27.11 & 1 & 37.36 & 0 & -     & 1 & 31.33 \\
3  & 1 & 29.07 & 1 & 30.92 & 1 & 31.83 & 1 & 30.94 \\
4  & 1 & 30.56 & 1 & 32.35 & 0 & -     & 1 & 32.33 \\
5  & 1 & 28.84 & 0 & -     & 0 & -     & 1 & 27.57 \\
6  & 0 & -     & 1 & 36.82 & 1 & 28.07 & 1 & 26.59 \\
7  & 0 & -     & 0 & -     & 1 & 27.85 & 0 & -     \\
8  & 1 & 29.81 & 0 & -     & 1 & 28.83 & 1 & 31.34 \\
9  & 0 & -     & 1 & 33.35 & 0 & -     & 0 & -     \\
10 & 1 & 28.05 & 1 & 32.37 & 0 & -     & 1 & 28.10 \\
\hline
Average time & \multicolumn{2}{c|}{\textbf{28.91}} & \multicolumn{2}{c|}{33.86} & \multicolumn{2}{c|}{29.15} & \multicolumn{2}{c}{29.35} \\
\hline
SR (success rate) & \multicolumn{2}{c|}{60\%} & \multicolumn{2}{c|}{60\%} & \multicolumn{2}{c|}{40\%} & \multicolumn{2}{c}{\textbf{80\%}} \\
\hline
\end{tabular}
\vspace{-1.0em}
\end{table}
\vspace{-1.0em}

The results indicate that both Gradient-Only and TTC-Only achieve a success rate of 60\%. 
However, the TTC-Only variant tends to produce repeated left-right oscillations, resulting in a much longer navigation time. The APF-Only method performs worse, with a success rate of only 40\%. 
In comparison, our method achieves the best overall performance, reaching an 80\% success rate while maintaining short navigation times through smooth trajectory outputs. 
Overall, the ablation results demonstrate that the proposed geometric planning module is more effective than each individual component alone.

\section{Conclusion}
In this paper, we proposed SemGeoNav, a hierarchical visual navigation framework for safe and robust goal-directed navigation. It combines high-level semantic reasoning with geometry-aware local planning. 
We conducted real-world experiments on the Unitree Go2 robot, and the results show that SemGeoNav outperforms representative baselines in success rate, navigation efficiency, and robustness. 
In the future, we aim to align the scale of depth observations with that of the waypoints predicted by the model to further improve navigation performance.

%
%

\end{document}